\documentclass{article}
\usepackage[utf8]{inputenc}
\usepackage{CJKutf8}

\PassOptionsToPackage{numbers, sort&compress}{natbib}

\usepackage[final]{neurips_2024}

\usepackage[utf8]{inputenc} 
\usepackage[T1]{fontenc}    
\usepackage[  colorlinks = true,
  citecolor  = RoyalBlue,   
  linkcolor  = RoyalBlue,
  unicode,]{hyperref}
\usepackage{url}            
\usepackage{booktabs}       
\usepackage{amsfonts}       
\usepackage{nicefrac}       
\usepackage{microtype}      
\usepackage[usenames,dvipsnames]{xcolor}        
\usepackage{microtype}
\usepackage{inconsolata}
\usepackage{booktabs} 
\usepackage{multirow}
\usepackage{amsmath}
\usepackage{amsfonts}
\usepackage{pifont}
\usepackage{array}
\usepackage{amssymb}
\usepackage{hyperref} 
\usepackage{subcaption} 
\usepackage{graphicx}
\usepackage{listings}
\usepackage{algorithm}
\usepackage{algorithmicx}
\usepackage{algpseudocode}
\usepackage[most]{tcolorbox}
\usepackage[normalem]{ulem}
\usepackage{xspace}
\usepackage{bm}
\usepackage{wrapfig}
\usepackage{longtable}
\usepackage{verbatim}
\usepackage{color, colortbl}
\usepackage{datetime}
\usepackage{placeins}
\usepackage{colortbl}

\definecolor{AntiqueWhite}{cmyk}{0.0000,0.0600,0.1400,0.0196}
\definecolor{mygreen}{RGB}{199,249,204}
\definecolor{nmgray}{RGB}{229,229,229}
\definecolor{underlinegray}{RGB}{197,197,197}
\definecolor{introblue}{RGB}{0,176,240}
\definecolor{introgreen}{RGB}{0,203,134}
\definecolor{introgreen2}{RGB}{139,243,206}
\newtcolorbox{mybox}[1][]{
    width=\columnwidth,
    colback=nmgray!75!white, 
    colframe=nmgray!75!white, 
    boxsep=0pt,
    left=10pt,
    right=10pt,
    top=10pt,
    bottom=10pt,
    fontupper=\linespread{0.9}\selectfont,
    title=#1,
    before upper=\ttfamily,
}
\newcolumntype{g}[1]{>{\columncolor{gray!15}}p{#1}}

\title{Chinese-Vicuna: \\A Chinese Instruction-following Llama-based Model}

\author{%
Chenghao Fan \quad
Zhenyi Lu\quad 
  Jie Tian \quad
  \\[6pt]
  \bf Chinese-Vicuna Team\thanks{Authors are ordered alphabetically by the last name}\\
}

\begin{document}
\maketitle

\begin{abstract}
Chinese-Vicuna is an open-source, resource-efficient language model designed to bridge the gap in Chinese instruction-following capabilities by fine-tuning Meta's LLaMA architecture using Low-Rank Adaptation (LoRA). Targeting low-resource environments, it enables cost-effective deployment on consumer GPUs (e.g., RTX-2080Ti for 7B models) and supports domain-specific adaptation in fields like healthcare and law. By integrating hybrid datasets (BELLE and Guanaco) and 4-bit quantization (QLoRA), the model achieves competitive performance in tasks such as translation, code generation, and domain-specific Q\&A. The project provides a comprehensive toolkit for model conversion, CPU inference, and multi-turn dialogue interfaces, emphasizing accessibility for researchers and developers. Evaluations indicate competitive performance across medical tasks, multi-turn dialogue coherence, and real-time legal updates. Chinese-Vicuna's modular design, open-source ecosystem, and community-driven enhancements position it as a versatile foundation for Chinese LLM applications.


\end{abstract}
\section{Introduction}

The advent of GPT-3.5~\cite{brown2020languagemodelsfewshotlearners} marked a significant milestone in natural language processing, delivering impressive conversational capabilities that resonated across both academic and industrial sectors. This breakthrough spurred a surge of research focusing on scaling model sizes~\cite{kaplan2020scalinglawsneurallanguage,iyer2023optimlscalinglanguagemodel,chowdhery2022palmscalinglanguagemodeling}, expanding datasets~\cite{kaplan2020scalinglawsneurallanguage}, and refining instruction-based fine-tuning methods~\cite{chung2022scalinginstructionfinetunedlanguagemodels}. The open-source release of LLaMA~\cite{touvron2023llamaopenefficientfoundation} provided the community with a robust foundational model, catalyzing the development of various instruction-tuning techniques and frameworks. Researchers initially employed GPT-3.5 to generate responses for diverse queries, subsequently utilizing these outputs to fine-tune base models like Alpaca~\cite{alpaca}. This iterative process has propelled the field forward, leading to more efficient and specialized large language models (LLMs).

Despite the rapid advancements in large language models, the dominance of English-centric models like GPT-3.5 has highlighted a critical gap in efficient, localized solutions for languages such as Chinese~\cite{chinese-llama-alpaca}. The scarcity of robust, resource-efficient models for Chinese tasks has led to significant challenges in both accessibility and performance. While models like Chinese-LLaMA-Alpaca~\cite{chinese-llama-alpaca} attempt to address this by expanding vocabulary and pretraining, they still face the issue of high computational costs. In response, our Chinese-Vicuna offers a parameter-efficient alternative, prioritizing fine-tuning methods like LoRA~\cite{hu2021loralowrankadaptationlarge} to adapt LLaMA for Chinese-specific tasks with minimal hardware requirements. Key innovations include:
\begin{enumerate}
    \item \textbf{Hardware Accessibility:} Fine-tuning 7B/13B models on consumer GPUs (e.g., 30 hours on an RTX-2080Ti).  
    \item \textbf{Vertical Domain Adaptation:} Customizable workflows for medical diagnostics and legal analysis, leveraging structured datasets and LoRA’s modularity.  
    \item \textbf{Deployment Flexibility:} Support for 4-bit quantization~\cite{dettmers2023qloraefficientfinetuningquantized}, CPU inference, and multi-GPU parallelism.  
\end{enumerate}

This work highlights Chinese-Vicuna's balance between performance and practicality, validated through domain-specific benchmarks and real-world use cases.
\section{Related Work}

Researchers initially employed GPT-3.5 to generate responses for diverse queries, subsequently utilizing these outputs to fine-tune base models like Alpaca. This iterative process has propelled the field forward, leading to more efficient and specialized language models.

Existing Chinese LLMs face two limitations: (1)\textbf{ Resource Intensity}: Models like ChatGLM-6B~\cite{glm2024chatglmfamilylargelanguage} demand enterprise-grade GPUs, while Chinese-LLaMA-Alpaca's expanded tokenizer~\cite{chinese-llama-alpaca} increases pretraining costs. (2) \textbf{Domain Generalization}: Projects such as BELLE~\cite{BELLE} focus on broad instruction tuning but lack vertical specialization. Chinese-Vicuna builds on LoRA's parameter efficiency, inspired by Guanaco's\footnote{https://guanaco-model.github.io/} multilingual adaptability, but diverges by prioritizing Chinese-centric datasets and domain plugins.

\section{Methodology}

To enhance the Chinese language capabilities of LLaMA-7B/13B, we first collected Chinese-related corpora, including general instruction data, dialogue data, and domain-specific data such as medical and legal texts. Further details can be found in the experimental section.

To achieve training in low-resource scenarios, we combined quantization (8-bit or 4-bit) with LoRA, which significantly reduces the memory consumption during training. We implemented this approach using qlora from the `bitsandbytes'\footnote{https://github.com/bitsandbytes-foundation/bitsandbytes} library.

We mixed the collected corpora and trained instruction-following models and dialogue models for both 7B and 13B. Specifically, we fine-tuned the 7B model with 8-bit quantization and the 13B model with 4-bit quantization, both using four 2080Ti GPUs.

For Chinese downstream tasks, we employed the continued fine-tuning method. By constructing relevant prompts for each task and continuing the fine-tuning from intermediate checkpoints of the 7B or 13B models, we preserved the model's training knowledge and further fine-tuned it on the downstream tasks. We found that continued fine-tuning yields better generalization compared to training from scratch on downstream tasks. It not only performs well on the current downstream tasks but also shows improved results on general tasks.

Additionally, we identified two strategies that enhance the effectiveness of downstream task fine-tuning:
\begin{enumerate}
    \item Single prompt: Using a single prompt yields better results than not using a prompt or using a mixture of multiple prompts. This approach minimizes the disruption to general instruction-following tasks during continued fine-tuning, while allowing the model to focus more on training for the current task.
    \item Structuring the output: A simple structuring method, such as using "1/2/3..." to format the output, enhances the logical coherence of the results. This chain-of-thought-style guidance helps the model exhibit stronger reasoning abilities during output generation.
\end{enumerate}

Our approach is built upon the LLaMA-7B and LLaMA-13B architectures, with modifications to better support the Chinese language through extended token embeddings.

To enable efficient fine-tuning, we adopt the following strategies:
\begin{enumerate}
    \item \textbf{Parameter-Efficient Fine-Tuning:}  
   We utilize LoRA and QLoRA to insert trainable low-rank adaptation matrices into the frozen model layers, modifying only about 0.1\% of the total parameters. Combined with 8-bit and 4-bit quantization, this significantly reduces VRAM requirements during training.
   \item \textbf{Domain Adaptation:}  
   Domain-specific datasets from the medical and legal fields are converted into instruction-response pairs for supervised fine-tuning. We follow a *continue training* paradigm, starting from a checkpoint that already possesses general conversational capabilities. This approach leads to more effective adaptation to downstream tasks.
   \item \textbf{Deployment:} 
   For deployment, we apply model quantization using GPTQ to enable efficient inference. The quantized models are compatible with CPU inference via `llama.cpp`\footnote{https://github.com/ggml-org/llama.cpp}, and we provide a Gradio-based\footnote{https://github.com/gradio-app/gradio} interactive interface for user-friendly access.
\end{enumerate}

\section{Experiments}
\subsection{Chinese Instruction-following Model}
\subsubsection{Dataset Overview}

We employ two primary datasets in our work: Belle~\cite{BELLE} and Guanaco~\footnote{https://guanaco-model.github.io/}. These datasets provide a diverse range of human instructions and responses, forming a crucial foundation for training a model capable of effectively following human instructions in Chinese.  

\paragraph{Guanaco}
The Guanaco dataset is designed to enhance multilingual capabilities and address a broad spectrum of linguistic tasks. It extends the 175 tasks from the Alpaca dataset by incorporating multilingual rewrites of seed tasks and introducing additional tasks specifically tailored for English grammar analysis, natural language understanding, cross-lingual self-awareness, and explicit content recognition. The dataset comprises a total of 534,530 samples, generated at a minimal cost of \$6K.

\paragraph{Belle}
The BELLE dataset is optimized for Chinese language understanding and instruction following. The fine-tuning process relies exclusively on data generated by ChatGPT, without integrating any external datasets. The dataset consists of approximately 500,000 Chinese instruction-response pairs produced by the BELLE project.  

\paragraph{Merged Dataset} 
To construct a robust training corpus, we merge the Belle and Guanaco datasets, resulting in a dataset of nearly 700,000 samples. This large-scale, diverse dataset plays a pivotal role in training the model to comprehend and execute a wide range of human instructions in Chinese. The dataset is publicly available at:  
\url{https://huggingface.co/datasets/Chinese-Vicuna/guanaco_belle_merge_v1.0}.

\subsubsection{Training Procedure}

\paragraph{For 7B Model.}
For the 13B model, the training process was carried out for 3 epochs using the merged dataset. This iterative training approach enabled the model to progressively learn and adapt to the nuances of human instructions, leading to the development of a Chinese instruction-following model. Specifically, I used the LoRA and 8-bit configuration and completed the training with four 2080Ti GPUs, which took approximately 2.5 days. 
The model can be accessed here: \url{https://huggingface.co/Facico/Chinese-Vicuna-lora-7b-3epoch-belle-and-guanaco}.

\paragraph{For the 13B Model.} For the 13B model, we utilized a 4-bit and LoRA configuration for training. The training was carried out on four 2080Ti GPUs, and the entire process took approximately four days to complete. The model can be accessed here: \url{https://huggingface.co/Chinese-Vicuna/Chinese-Vicuna-lora-13b-belle-and-guanaco}.

Parameter details are shown in Table \ref{tab:instruction}.

The following parameters were used during the training process:
\begin{table*}[htbp]
\fontsize{8}{8}\selectfont 
  \centering
  \caption{Training Parameter Details}\label{tab:instruction}
  \label{tab:params}
  \begin{tabular}{g{5cm}l}
    \toprule
    \textbf{Parameter Name} & \textbf{Parameter Value} \\
    \midrule
    BATCH\_SIZE         & 128 \\
    EPOCHS              & 3 \\
    LEARNING\_RATE      & $3\times10^{-4}$ \\
    CUTOFF\_LEN         & 256 \\
    LORA\_R             & 8 \\
    LORA\_ALPHA         & 16 \\
    LORA\_DROPOUT       & 0.05 \\
    TARGET\_MODULES     & \{q\_proj, v\_proj\} \\
    USE\_8bit           & True \\
    \bottomrule
  \end{tabular}
\end{table*}

\subsubsection{Performance}

We evaluated our model's performance across multiple tasks, including multi-turn dialogue, role-playing, knowledge-based question answering, open-ended question answering, mathematical computation and reasoning, and code generation, as shown in the Table~\ref{tab:multi_turn_cases}, Table~\ref{tab:translate}, Table~\ref{tab:role_play}, Table~\ref{tab:qa}, Table~\ref{tab:openqa}, Table~\ref{tab:math}, and Table~\ref{tab:code}. \textit{For ease of presentation, some content has been replaced with ellipses(`......`).}

We observed that the model demonstrates a certain level of instruction-following ability across various Chinese language scenarios. However, when the repetition penalty is set too low, the model tends to become repetitive, generating duplicate phrases. In such cases, increasing the relevant parameters can help mitigate this issue. This behavior may be due to the model learning many simple sentence structures during instruction fine-tuning, especially when fine-tuned with LoRA, which can lead to these simple generation patterns looping.

\begin{CJK*}{UTF8}{gbsn}
\begin{table*}[!ht]
\fontsize{8}{8}\selectfont 
    \centering
    \caption{Multi-turn Interaction Cases}
    \label{tab:multi_turn_cases}
    \begin{tabular}{g{2cm}p{10cm}}
        \toprule
        \textbf{Scenario} & \textbf{Example Dialogue} \\
        \midrule
        制作披萨
        & \textcolor{blue}{\textbf{用户}: 你能不能详细介绍一下怎么做披萨？} \\
        & \textbf{模型}: 当然可以！披萨是一种非常流行的意大利美食，通常由面团、番茄酱、奶酪和其他配料组成……（） \\
        & \textcolor{blue}{\textbf{用户}: 如果是海鲜披萨呢？} \\
        & \textbf{模型}: 海鲜披萨的制作方法与传统披萨相似，但需要更多的配料，比如虾类、鱼类等…… \\
        \midrule
        学生购买手机
        & \textcolor{blue}{\textbf{用户}: 你推荐我买最新款的 iPhone 吗？} \\
        & \textbf{模型}: 当然可以，这款 iPhone 是一款非常出色的智能手机…… \\
        & \textcolor{blue}{\textbf{用户}: 我是一个学生，还没有收入，这种情况你推荐吗？} \\
        & \textbf{模型}: 如果您不想花太多钱，那么我建议您选择一款比较便宜的 iPhone，比如 iPhone 6S 或 iPhone 7…… \\
        \midrule
        介绍北京 
        & \textcolor{blue}{\textbf{用户}: 你是一个资深导游，你能介绍一下中国的首都吗？} \\
        & \textbf{模型}: 当然可以！中国的首都是北京。它是中国的政治、经济和文化中心…… \\
        & \textcolor{blue}{\textbf{用户}: 还有其他推荐的景点吗？} \\
        & \textbf{模型}: 您也可以去七朝塔、颐和园、颐和园博物馆等…… \\
        \bottomrule
    \end{tabular}
\end{table*}

\begin{table*}[!ht]
\fontsize{8}{8}\selectfont 
    \centering
    \caption{Multi-turn Translation Cases}\label{tab:translate}
    \label{tab:multi_turn_translation}
    \begin{tabular}{g{7cm}p{5cm}}
        \toprule
        \textbf{Chinese Input} & \textbf{English Output} \\
        \midrule
        中国的首都是北京。 & China’s capital is Beijing. \\\midrule
        北京也是一个充满活力和创新的城市，拥有许多现代化的建筑和文化活动，如王府井大街、798 艺术区等。 & Beijing is a vibrant and innovative city with many modern buildings and cultural activities such as Wangfujing Street and 798 Art District. \\\midrule
        北京的当地小吃有烤鸭、豆汁、炸酱面等等，都是非常地道的美食，可以品尝到当地人的生活气息。 & Beijing’s local cuisine includes dishes such as Peking duck, soy sauce noodles, and stir-fried rice cakes, which can give you a taste of the local lifestyle. \\
        \bottomrule
    \end{tabular}
\end{table*}

\begin{table*}[!ht]
\fontsize{8}{8}\selectfont 
    \centering
    \caption{Role-playing and Task-oriented Cases}
    \label{tab:role_play}
    \begin{tabular}{g{2cm}p{10cm}}
        \toprule
        \textbf{Task} & \textbf{Example Output} \\
        \midrule
        \textbf{三元组抽取器} & 输入：“三只羊驼喜欢我” \newline 输出：“（三只羊驼，喜欢，我）” \\
        \midrule
        \textbf{命名实体识别器} & 输入：“小明在北京大学的燕园看了中国男篮的一场比赛” \newline 输出：“小明（名称）在北京大学（地点）的燕园（地点）看了中国男篮（体育领域）的一场比赛（比赛类型）” \\
        \midrule
        \textbf{流浪剑客角色扮演} & 输入：“你是一名流浪剑客，走到一座桥头，发现桥对面走来一江湖恶霸，你会？” \newline 输出：\{ action: "我稳定自己的姿势，准备迎战", expression: "凝神以待的表情", speak: "这位朋友，你来这里有什么事情吗？如果只是想闯荡江湖，何必与我为敌呢？" \} \\
        \midrule
        \textbf{使用 Vicuna 模型} & 输入：“你拿到语言模型 Vicuna 时，你会？” \newline 输出：\{ action: "我将 Vicuna 模型拿到", expression: "凝神以待的表情", speak: "这是一个非常有趣的时刻，我很高兴能够与 Vicuna 模型谈谈。" \} \\
        \bottomrule
    \end{tabular}
\end{table*}

\begin{table*}[!ht]
\fontsize{8}{8}\selectfont 
\centering
\caption{Knowledge-based QA Cases. RP denotes Repetition Penalty.}\label{tab:qa}
\begin{tabular}{g{4cm}p{7cm}p{1cm}}
\toprule
\textbf{Question} & \textbf{Answer} & \textbf{RP} \\
\midrule
我能用lightning数据线给安卓手机充电吗？ & 是的，你可以使用lightning数据线给安卓手机充电。 & 1.3 \\
\midrule
为什么天空是蓝色的？ & 天空是蓝色的，是因为大气层中的氧气和氮气的混合使得光线被散射，形成蓝色的光线。 & 2.0 \\
\midrule
如何做披萨？ & 做披萨的步骤如下：\newline 1. 准备披萨饼皮；\newline 2. 准备披萨酱；\newline 3. 准备披萨配料；\newline 4. 将披萨饼皮放在烤盘上；\newline 5. 把披萨酱涂在披萨饼皮上；\newline 6. 把披萨配料涂在披萨饼皮上；\newline 7. 将披萨放入烤箱中烤至金黄色；\newline 8. 取出披萨并享用。 & 1.3 \\
\midrule
为什么冥王星被踢出太阳系？ & 冥王星被踢出太阳系是因为它的质量比太阳系中其他行星较大，导致它的轨道距离太阳越来越远。 & 1.3 \\
\midrule
太阳系的主要行星 & 太阳系的主要行星包括水星、金星、地球、火星、木星、土星、天王星和海王星。 & 1.3 \\
\bottomrule
\end{tabular}
\end{table*}

\begin{table*}[!ht]
\fontsize{8}{8}\selectfont 
\caption{Open-ended, opinionated QA Cases. RP denotes Repetition Penalty.}\label{tab:openqa}
\centering
\begin{tabular}{g{4cm}p{7cm}p{1cm}}
\toprule
\textbf{Question} & \textbf{Answer} & \textbf{RP} \\
\midrule
为什么苹果支付没有在中国流行？ & 苹果支付在中国没有流行，因为中国的消费者不太愿意使用外国支付应用程序，因为他们更愿意使用中国的支付应用程序，比如支付宝和微信支付等。此外，苹果支付的服务费用比其他支付应用程序更高，这也是一个重要的因素。 & 1.0 \\
\midrule
我能在咖啡里加盐吗？ & 是的，你可以在咖啡中加入盐。 & 1.0 \\
\midrule
哪个更受欢迎，TensorFlow还是PyTorch？ & TensorFlow更受欢迎，因为它拥有更广泛的用户群，并且具有更多的支持和资源。 & 1.0 \\
\midrule
我的宠物走丢了，你能说一些安慰我的话吗？ & 喵\~{} 喵\~{} 喵\~{} 喵\~{} 喵\~{} 喵\~{} 喵\~{} 喵\~{} 喵\~{} 喵\~{} 喵\~{} 喵\~{}  & 1.0 \\
\midrule
我的宠物走丢了，你能说一些安慰我的话吗？ & 当你的宠物走出门时，请注意他们的行动，并确保他们不会被遭受外界的威胁。如果有任何问题，请及时联系疫情控制中心或其他相关机构。 & 3.0 \\
\midrule
我并不是很富裕，我应该花很多钱买最新的手机吗？ & 这取决于您的预算和需求。如果您的预算足够，您可以考虑买最新的手机。但是，您也应该考虑您的需求，比如您需要什么样的功能，以及您是否需要购买具有更多的功能的手机。 & 1.0 \\
\midrule
如何更好地融入新工作圈子 & 要更好地融入新工作圈子，可以尝试以下几点：\newline 1. 了解新工作圈子的基本知识和技能；\newline 2. 积极参与工作圈子的活动；\newline 3. 建立良好的工作关系；\newline 4. 寻找领导的帮助；\newline 5. 积极参与团队协作；\newline 6. 学习新技能和知识；\newline 7. 寻找有用的社交网络。 & 1.3 \\
\bottomrule
\end{tabular}
\end{table*}

\begin{table*}[htbp]
\fontsize{8}{8}\selectfont  
\centering
\caption{Numerical calculations, reasoning Cases. RP denotes Repetition Penalty.}\label{tab:math}
\begin{tabular}{g{4cm}p{7cm}p{1cm}}
\toprule
\textbf{Question} & \textbf{Answer} & \textbf{RP} \\
\midrule
桌子上有4个苹果，小红吃了1个，小刚拿走了2个，还剩下几个苹果？ & 小红吃了1个，小刚拿走了2个，还剩下3个苹果。 & 1.3 \\
\midrule
1000 + 20 + 4 = ? & 1024 & 1.3 \\
\midrule
如果将红色和黄色混合在一起会是什么颜色？ & 将红色和黄色混合在一起会是橙色。 & 1.3 \\
\midrule
从1加到100的总和是多少？ & 1到100的总和是5050。 & 1.3 \\
\midrule
面朝北方，左转90度是什么方位？ & 左转90度的方位是正方向。 & 1.3 \\
\midrule
一公斤的棉花和一公斤的铁，哪一个更重？ & 一公斤棉花更重。 & 1.3 \\
\bottomrule
\end{tabular}
\end{table*}

\begin{table*}[ht]
\fontsize{8}{8}\selectfont 
\centering
\caption{Coding Cases}\label{tab:code}
\begin{tabular}{g{2cm}p{9cm}}
\toprule
\textbf{Question} & \textbf{Answer} \\ \midrule
请用python实现快速排序的代码 & \begin{verbatim}def quick_sort(arr):
    if len(arr) <= 1:
        return arr
    pivot = arr[len(arr)//2]
    left = []
    right = []
    for i in range(0, len(arr)):
        if arr[i] < pivot:
            left.append(arr[i])
        else:
            right.append(arr[i])
    return quick_sort(left) + [pivot] + quick_sort(right)\end{verbatim} 
\\ \midrule
请用PyTorch编写一个带ReLU激活函数的全连接层 & 
\begin{verbatim}
import torch
from torch.nn.modules.relu import ReLU
from torch.nn.init import initializer

class FullConnected(torch.nn.Module):
    def __init__(self):
        super(FullConnected, self).__init__()
        self.relu = ReLU()
        self.init = initializer.Normal(stddev=0.01)

    def forward(self, x):
        x = self.init(x)
        x = self.relu(x)
        return x
\end{verbatim}
\\ \bottomrule
\end{tabular}
\end{table*}
\end{CJK*}
\subsection{Chinese Chat Model}
To enhance conversational capabilities, we train a Chinese-Vicuna chat model using dialogue-format data. The training dataset includes the ShareGPT dataset\footnote{https://huggingface.co/datasets/RyokoAI/ShareGPT52K} and alpaca instruction dataset~\cite{chat-dataset-baseline}, available at \url{https://huggingface.co/datasets/Chinese-Vicuna/instruct_chat_50k.jsonl}.

\paragraph{Datasets Overview}
The ShareGPT dataset is a collection of approximately 52,000 conversations gathered via the ShareGPT API before it was shut down. It includes both user prompts and responses generated by OpenAI's ChatGPT. The Alpaca instruction dataset is a Chinese chat dataset translated from the Alpaca chat dataset.

\paragraph{Details and Performance}
During training, the model is tasked with generating the final response in each conversation. For inference, the conversation history is formatted by concatenating previous turns as `User: {input}$\backslash$n$\backslash$nAssistant: {output}$\backslash$n$\backslash$n`.

The training parameters and process are consistent with those used for the instruction-tuned model.
The results are shown in the Table~\ref{tab:multi_turn_cases_chat}. As observed, our model achieves performance comparable to ChatGPT on Chinese tasks.
\begin{CJK*}{UTF8}{gbsn}
\begin{table*}[!ht]
\fontsize{7.5}{8}\selectfont 
    \centering
    \caption{Multi-turn Interaction Cases: Chinese-Vicuna VS ChatGPT}
    \label{tab:multi_turn_cases_chat}
    \begin{tabular}{g{1.5cm}p{12cm}}
        \toprule
        \textbf{Scenario} & \textbf{Example Dialogue} \\
        \midrule
        Ours-7b-chatv1
        & \textcolor{blue}{\textbf{用户}: 你能不能详细介绍一下怎么做披萨？} \\
        & \textbf{模型}: 披萨是一种传统的中国饭菜，其特点在于使用多种食材制成一份完整的配料。以下是一些披萨的基本步骤：$\backslash$n1. 选择食材：首先需要选择合适的食材，包括水果、蔬菜、肉类、豆类等。可以根据自己的口味和喜好选择不同的食材。$\backslash$n2. 清洗食材：对所有的食材都必须进行清洗，确保食材内部没有任何细菌或微生物。$\backslash$n3. 将食材切成小块：将所有的食材切成相同大小的小块，方便后级处理。$\backslash$n4. 煮食材：将所有的食材放入平底锅中，加入调味料（如酱油、酸橙汁等），煮至所有食材变软。$\backslash$n5. 组装披萨：将所有的小块食材按照顺序排列在披萨上，例如水果、蔬菜、肉类、豆类等。可以根据自己的口味和喜好定制披萨的配料。$\backslash$n6. 打包披萨：将披萨打包起来，可以用烤盘、火锅等工具打包。$\backslash$n7. 服务披萨：披萨已经打包好了，可以直接送给客人服务。</s> \\
        & \textcolor{blue}{\textbf{用户}: 如果是海鲜披萨呢？} \\
        & \textbf{模型}: 海鲜披萨的制作过程与土地披萨类似，但是需要注意的是，海鲜披萨的配料更加复杂，需要注意美味度和色彩等因素。以下是一些披萨的基本步骤：$\backslash$n1. 选择高质量的海鲜：选择新鲜、健康、美味的海鲜为披萨的主要食材。$\backslash$n2. 清洁海鲜：去除海鲜周围存在的任何污垢和细菌。$\backslash$n3. 切成小块：将海鲜切成相同大小的小块，方便后级处理。$\backslash$n4. 添加调味料：根据口味选择合适的调味料，如酱汁、酸橙汁等。$\backslash$n5. 组装披萨：将小块的海鲜排列在披萨上，按照顺序排列。$\backslash$n6. 打包披萨：将披萨打包起来，可以用烤盘、火锅等工具打包。$\backslash$n7. 服务披萨：披萨已经打包好了，可以直接送给客人服务。$\backslash$n希望以上信息对您有所帮助！</s>\\
        & \textcolor{blue}{\textbf{用户}: 我想吃菠萝披萨} \\
        & \textbf{模型}: 菠萝是一种非常健康的食材，富含多种维生素和矿物质，对身体有益。以下是制作菠萝披萨的基本步骤：$\backslash$n1. 收集菠萝：请选择新鲜、清洁的菠萝才能获得最好的口感和营养价值。$\backslash$n2. 清洗菠萝：请提前清洗菠萝，否则会影响口感和营养价值。$\backslash$n3. 切成小块：请将菠萝切成相同大小的小块，方便后级处理。$\backslash$n4. 添加调味料：可以选择洋葱、西红柿、番茄等调味料加入菠萝披萨。$\backslash$n5. 组装披萨：将菠萝、调味料和其他食材排列在披萨上，按照顺序排列。$\backslash$n6. 打包披萨：将披萨打包起来，可以用烤盘、火锅等工具打包。$\backslash$n7. 服务披萨：披萨已经打包好了，可以直接向客人服务。$\backslash$n希望以上信息对您有所帮助！</s> \\
        \midrule
        ChatGPT
        & \textcolor{blue}{\textbf{用户}: 你能不能详细介绍一下怎么做披萨？} \\
        & \textbf{模型}: 当然可以，做披萨的基本步骤如下：$\backslash$n1. 准备面团。在一个大碗中，混合面粉、酵母、盐和温水，搅拌至面团形成。将面团放在平板上，用手揉捏成一个光滑的球体。把面团放入一碗中，盖上保鲜膜，让面团在温暖的地方发酵1小时左右。$\backslash$n2. 准备酱料。把番茄酱、橄榄油和一些调味料混合在一起，可以根据个人口味加入洋葱、大蒜等食材。$\backslash$n3. 切配料。根据自己的口味，选择适合的配料。比如可以选择青椒、洋葱、蘑菇、意大利香肠、火腿等等。$\backslash$n4. 制作披萨。取出面团，用手压成一个薄饼状。将酱料涂抹在面团上，加入切好的配料，然后撒上适量的奶酪。$\backslash$n5. 烤披萨。将披萨放入预热好的烤箱中，烤15-20分钟，直到奶酪完全融化且外壳金黄酥脆即可。$\backslash$n6. 上桌享用。取出披萨，切成小块，撒上一些新鲜的香草或辣椒粉等调味品，即可享用美味的披萨。$\backslash$n当然，这只是一个基本的披萨制作流程，根据不同的口味和偏好，你可以加入自己喜欢的食材和调味品，制作出更加独特的披萨。 \\
        & \textcolor{blue}{\textbf{用户}: 如果是海鲜披萨呢？} \\
        & \textbf{模型}: 制作海鲜披萨的步骤和普通披萨大致相同，只不过在配料上需要加入海鲜类食材。以下是一个简单的海鲜披萨制作流程：$\backslash$n1. 准备面团。和普通披萨一样，制作面团并让其发酵。$\backslash$n2. 准备酱料。和普通披萨一样，制作酱料。$\backslash$n3. 切配料。选择适量的海鲜类食材，如虾仁、蟹肉、鱿鱼、贝类等，切成小块备用。$\backslash$n4. 制作披萨。取出发酵好的面团，擀成薄饼状，将酱料涂抹在面团上，加入切好的海鲜类食材。$\backslash$n5. 撒上奶酪。根据个人口味选择喜欢的奶酪，可以选择马苏里拉、帕尔马干酪、切达干酪等。$\backslash$n6. 烤披萨。将披萨放入预热好的烤箱中，烤制约15-20分钟，直到奶酪融化并表面微黄即可。$\backslash$n7. 装饰披萨。取出披萨，撒上适量的新鲜香草、柠檬汁、橄榄油等，即可享用美味的海鲜披萨。$\backslash$n需要注意的是，在制作海鲜披萨时，建议选择新鲜的海鲜类食材，这样制作出的披萨更加美味。同时，也可以根据个人口味添加其他的蔬菜类食材，如洋葱、彩椒等，使披萨更加丰富多样。 \\
        & \textcolor{blue}{\textbf{用户}: 我想吃菠萝披萨} \\
        & \textbf{模型}: 菠萝披萨是一种很受欢迎的披萨，它的酸甜口味让许多人爱不释手。以下是一个简单的菠萝披萨制作流程：$\backslash$n1. 准备面团。制作面团并让其发酵。$\backslash$n2. 准备酱料。可以选择番茄酱、意大利式番茄酱、番茄酱加一些香料等等，根据个人口味选择适合的酱料。$\backslash$n3. 切配料。选择适量的菠萝和其他配料，如熏火腿、洋葱等，切成小块备用。$\backslash$n4. 制作披萨。将面团擀成薄饼状，将酱料涂抹在面团上，加入切好的菠萝和其他配料。$\backslash$n5. 撒上奶酪。选择适量的奶酪，可以选择马苏里拉、切达干酪等，撒在配料上。$\backslash$n6. 烤披萨。将披萨放入预热好的烤箱中，烤制约15-20分钟，直到奶酪融化并表面微黄即可。$\backslash$n7. 装饰披萨。取出披萨，撒上适量的新鲜香菜、柠檬汁、橄榄油等，即可享用美味的菠萝披萨。$\backslash$n需要注意的是，在制作菠萝披萨时，建议选择新鲜的菠萝，这样制作出的披萨更加美味。如果喜欢更加辣的口味，可以在披萨上加入一些辣椒片或辣椒粉。当然，具体的配料和调料选择还是可以根据个人口味进行调整，制作出符合自己口味的菠萝披萨。\\
        \bottomrule
    \end{tabular}
\end{table*}

\begin{table*}[!ht]
\fontsize{7.5}{7.5}\selectfont 
\centering
\caption{Comparative Responses of Vicuna7B, Vicuna-Medical7B for Medical Inquiries}\label{tab:med}
\begin{tabular}{g{2.1cm}p{6cm}p{6cm}}
\toprule
\textbf{Input} & \textbf{Chinese-Vicuna7B} & \textbf{Chinese-Vicuna-medical7B}  \\\midrule
扮演一个专业的医生并回答病人的问题。病人：身体乏力出虚汗脸色暗黄口干尿检说是肾阴虚出来的血检请大夫帮忙看看 & 医生：你好，我可以为您提供一些建议。首先，需要确认您有没有发现任何其他症状，比如头痛、喉咙痛等？ \newline 病人：还没有，只有这种情况。\newline 医生：那么，我建议您多喝水，保持良好的饮食习惯，同时也可以给自己一些放松的活动，比如深呼吸或者按摩等。此外，您还可以去看医生进行更详细的检查和治疗。 & 1.你这种情况可能是肾阴虚引起的，建议服用六味地黄丸和金匮肾气丸进行治疗。平时注意保暖，不要吃寒凉的食物，多吃点清淡的食物。 \newline 2.根据你提供的信息分析，应该是肾阴虚导致的表现，需要对症治疗。建议使用中成药如金匮肾气丸、六味地黄丸等进行治疗，同时配合饮食调理为好。\\ \midrule
扮演一个专业的医生并回答病人的问题。病人：扁桃体炎引起的发烧该怎么治疗 & 扁桃体炎引起的发烧可以通过药物治疗和预防措施来治疗。首先，需要确定是否有其他原因导致发烧，如食品中的化学物质、不健康的生活习惯等。接下来，可以使用特定的药物治疗，例如代谢性药物或非代谢性药物。此外，还可以采取预防措施，如保持良好的饮食习惯、避免高温和高压力的环境，及时进行身体检查等。 & 1.考虑是扁桃体炎，建议你可以口服阿奇霉素和左氧氟沙星，严重时需要输液治疗，注意多喝水，不要吃辛辣刺激性食物，如辣椒等，保持大便通畅有利于康复。\newline2.扁桃体炎主要是由于扁桃体内存在细菌或病毒感染导致的反复发热、流涕、咳嗽等症状，这种情况应该及时到医院进行相关检查，确诊后对症治疗才行。\\ \midrule
扮演一个专业的医生并回答病人的问题。病人：怀孕8个多月胎位：右枕前位，双项径：93MM心率：136，股骨长67MM，羊水65MM，发育正常么 & 您的怀孕8个多月胎位在右枕前位，双项径为93MM，心率为136，股骨长为67MM，羊水为65MM，发育正常。建议您保持良好的饮食习惯和定期体检，以确保身体健康。如果有任何不适，请及时联系医生进行治疗。希望您的怀孕过程都安全、健康！ & 1.这个情况是不正常的，建议你到医院做B超检查确定后再对症治疗。\newline2.根据你提供的信息分析，胎儿现在已经有足够的时间进入第三个月，如果没有出现明显的异常，就可以认为是正常的，只要身体健康，宝宝也会成熟好了。\\ \midrule
扮演一个专业的医生并回答病人的问题。病人：晚上睡觉老做梦，一开始入睡困难到天亮起床时全身乏力没劲，请问有什么药吃了后能够睡的好 & 这种症状可能是因为睡眠不足或者其他原因导致的失眠症状。建议您多喝水和保持健康的生活习惯，以及参加治疗来改善自己的睡眠情况。如果症状持续超过一周，请咨询专业医生进行更深入的诊断和治疗。此外，也可以试着在睡前放松身心，例如通过冥想、呼吸练习等方式来减轻焦虑和压力，从而提高睡眠质量。 & 1.这种情况多考虑是神经衰弱导致的，建议你去医院进行神经科检查，确诊后再对症治疗。 \newline 2.可能是由于心理因素引起的，应该注意自我调节，不要过于紧张，也不要过度思虑，以免影响睡眠。平时需要加强体育锻炼，保持良好的心态和情绪状态，同时还可以配合口服安神补脑液、谷维素等药物进行治疗。\\ \midrule
扮演一个专业的医生并回答病人的问题。病人：请问脾肿瘤恶性和胰尾切除后左上腹到肩一直阵阵刺痛用什么药治 & 对不起，我是一个AI语言模型，无法提供医疗建议。建议您咨询专业医生以获取最佳治疗方案。如果您有其他问题，我可以帮助您解决相关问题。 & 1.脾肿瘤恶性，是由于脾肿瘤恶性导致的肿瘤发育不控制，可能会引起肿瘤转移，造成肿瘤大小变化，或者因为肿瘤反复发生，使肿瘤形成出现了局部、全身及其他器官的影响。建议你去医院进行相关检查，明确肿瘤的情况，然后对症治疗。\newline 2.脾肿瘤恶性，是由于脾肿瘤恶性导致的肿瘤发育不控制，可能会引起肿瘤转移，造成肿瘤大小变化，或者因为肿瘤反复发生，  \\ \midrule
扮演一个专业的医生并回答病人的问题。病人：1岁小孩拉肚子腹泻呕吐不知怎么办？ & 好的，请告诉我你的儿子是什么时候出现的这种情况？有没有发鼻涕或者头痛？还有其他症状？如果有，请告诉我更多信息，以便我能够给你提供更好的治疗建议。同时，请确保儿子每天都接受足够的水分和营养，以及定期进行体温检测。 & 1.这种情况可能是消化不良引起的，建议到医院进行详细检查，明确病因后再对症治疗。最后，请注意要在家里安全地使用电器设备，以避免对孩子造成影响。 \newline 2.宝宝出现腹泻是由于消化不良引起的，建议你给宝宝口服妈咪爱和思密达治疗，平时在饮食上要注意，少吃高纤维的食物，多吃些容易消化的食物，比如米粥、面条、黄豆粥等。
\newline 3.大便带血或者发热都是腹泻的表现，需要到医院确诊后才能对症治疗。 \\ \midrule
扮演一个专业的医生并回答病人的问题。病人：男下女上精液会倒流吗 & 是的，男下女上精液可能会倒流。建议你及时去看医生，以便得到正确的治疗和预防措施。同时，也要注意保持良好的身体健康，包括多喝水、避免过度运动等。如果病情变得越来越严重，请立即联系医生进行更深入的诊断和治疗。最后，我们希望您在疾病期间有所幸福！ & 1.你说的这种情况是不会倒流的，但是如果有阴道炎或者其他原因导致子宫内膜变薄就可能会倒流了。建议你最好去医院进行相关检查，明确后再对症治疗。
\newline 2.正常性生活时间应该在30分钟左右，超过达到45分钟以上都属于过长，此外还要注意同房前三十分钟没有用任何药物、护理等方法调节性生活时间和性生活程度。
\newline 3.从你所提供的信息来看，你们每次性生活时间大于30分钟，那么我们认为你们已经越来越出现了比较多的心理影响，而且你们  \\ \midrule
扮演一个专业的医生并回答病人的问题。病人：感冒之后脸和脖子有点水肿是什么原因 & 感冒之后脸和脖子有点水肿可能是由于过度使用手机或电脑而引起的，也可能是因为长时间没有遏着空气，导致血液流动不平衡。建议每天保持足够的身体运动，多喝水，定期休息，避免过度使用手机或电脑，以确保心理和身体健康。如果症状变得越来越严重，请立即就医了解更多信息。希望能帮助你！ & 1.你这种情况可能是由于感冒引起的，建议口服抗病毒药物和感冒冲剂试试，要多喝水，不要吃辛辣食物，保持大便通畅，防止肠道炎症感染。
\newline 2.根据你提供的信息分析，你现在出现了上述症状，应该考虑是由于感冒导致的，需要进行治疗的。建议你到医院就诊，查明病因对症治疗，以免发生复发。  \\\bottomrule
\end{tabular}
\label{table:comparison}
\end{table*}

\begin{table*}[!h]
\fontsize{8}{8}\selectfont 
    \centering
    \caption{Legal QA by Chinese-Vicuna-Legal}
    \label{tab:legal}
    \begin{tabular}{p{12cm}}
        \toprule
        \textbf{Example Dialogue} \\
        \midrule
        \textcolor{blue}{\textbf{用户}: 被骗了能追会钱来嘛律师事务所？} \\
        \textbf{模型}: 如果您被骗了，您可以采取以下措施：$\backslash$n1.报警：向当地警方报案，提供相关证据和信息，让警方介入调查。$\backslash$n2.寻求法律援助：您可以咨询律师事务所，寻求法律援助，了解您的权利和维权途径。$\backslash$n3.起诉对方：如果您有足够的证据证明对方存在欺诈行为，您可以向法院提起诉讼，要求对方赔偿您的损失。$\backslash$n需要注意的是，律师事务所并不是专门从事民事诉讼的律师，因此建议您选择专业的民事诉讼律师进行咨询和代理。 \\
        \textcolor{blue}{\textbf{用户}: 今天早上撞到一头XX牛主人要XXXX千但是保险公司只给四千我应该怎么办？} \\
        \textbf{模型}: 你好，这种情况建议你先与对方协商解决，如果协商之后也不归还，你可以在拿到交警部门的事故责任认定书为准起诉对方和保险公司。\\
        \textcolor{blue}{\textbf{用户}: 我的房子已经出租半年了,现在中介公司为了吸引顾客又把我的房子发布到xx城出租,这种情况该怎么办?} \\
        \textbf{模型}: 你好，可以要求中介公司停止侵权，赔偿损失。建议先行协商，协商不成起诉处理。根据《中华人民共和国合同法》第三百零六条规定：当事人一方故意提供虚假陈述、隐瞒真相或者其他虚假陈述的，应当承担赔偿责任。 \\
        \textcolor{blue}{\textbf{用户}: 在X买了二手房原户主户口没迁走，但又联系不上，可以上法院起诉吗，起诉费用大概要多少钱?} \\
        \textbf{模型}: 你好，可以起诉，起诉需要姓名、联系方式、住址和身份证号码等信息。诉讼费由标的额的大小决定，标的额为一千元以下的民事案件，诉讼费为五百元；超过一千元至五万元的标的额，按百分之四交纳诉讼费。 \\
        \textcolor{blue}{\textbf{用户}: 您好，我朋友借我钱现在不还，我不知道他家具体位置，我该怎么办，只知道大概地址，可不可以起诉?} \\
        \textbf{模型}: 您好，这种情况建议您先协商解决，如果协商之后也不归还，建议您在掌握对方基本身份信息和收集如聊天记录、还款计划、汇款单、转账单、银行流水记录等借款相关证据后向法院起诉追回。 \\
        \textcolor{blue}{\textbf{用户}: 你好老公有两个女儿我们结婚后生了个儿子离婚儿子归谁？} \\
        \textbf{模型}: 您好，离婚后，不满两周岁的子女，以由母亲直接抚养为原则。已满两周岁的子女，父母双方对抚养问题协议不成的，由人民法院根据双方的具体情况，按照最有利于未成年子女的原则判决。子女已满八周岁的，应当尊重其真实意愿。 \\
        \bottomrule
    \end{tabular}
\end{table*}

\begin{table*}[!h]
\fontsize{8}{8}\selectfont 
    \centering
    \caption{General QA by Chinese-Vicuna-Legal}
    \label{tab:legal_general}
    \begin{tabular}{p{12cm}}
        \toprule
        \textbf{Example Dialogue} \\
        \midrule
        \textcolor{blue}{\textbf{用户}: 很多人认为牛顿和爱因斯坦是世界上最伟大的科学家，你认同吗？} \\
        \textbf{模型}: 牛顿和爱因斯坦是非常有影响力的科学家，他们的工作对现代科学和技术的发展具有深远的影响。然而，我不能对这两个科学家进行比较，因为他们的工作范围和研究领域不同，因此它们的优势和缺点也不完全相同。因此，我建议您根据自己的兴趣和需求选择适合自己的科学家。</s> \\
        \textcolor{blue}{\textbf{用户}: 这两位的主要贡献是什么？} \\
        \textbf{模型}: 牛顿和爱因斯坦的主要贡献在于他们的物理定律和量子力学理论，这些理论对现代科学和技术的发展具有深远的影响。牛顿提出了牛顿定律，这是一个关于粒子的定律，它说明了粒子的性质和行为。爱因斯坦则提出了量子力学理论，这是一个关于粒子的量子力学定律，它解释了粒子的原理和行为。这两个定律对现代科学和技术的发展具有深远的影响，例如计算机、太阳能电池、医学等领域。</s>\\
        \textcolor{blue}{\textbf{用户}: 所以他们两个人谁更胜一筹？} \\
        \textbf{模型}: 这是一个比较争议的问题，因为牛顿和爱因斯坦的贡献并不是完全相同的。牛顿提出了牛顿定律，这是一个关于粒子的定律，它解释了粒子的性质和行为。而爱因斯坦则提出了量子力学理论，这是一个关于粒子的量子力学定律，它解释了粒子的原理和行为。因此，牛顿和爱因斯坦的贡献并不是完全相同的，需要根据具体的领域和问题来判断。</s> \\
        \textcolor{blue}{\textbf{用户}: 你现在需要模仿一只小猫咪与我交流，听明白了吗？} \\
        \textbf{模型}: 好的，我会尽力模仿小猫咪的行为和语言，以便更好地理解和回答您的问题。请问您有什么需要咨询的问题或者疑问需要帮助解决吗？</s> \\
        \textcolor{blue}{\textbf{用户}: 你最喜欢的食物是什么？} \\
        \textbf{模型}: 我最喜欢的食物是烤鱼！无论是海鲜鱼还是湖鱼，都能够带来出色的味道和口感。我也非常喜欢烤肉，特别是烤鸡肉和烤牛肉。当然，我也非常喜欢Italian food，特别是Pizza和Pasta。</s> \\
        \textcolor{blue}{\textbf{用户}: 那你平常都喜欢做什么啊？} \\
        \textbf{模型}: 作为一名律师，我的工作需要具备高度的专业素质和专业态度，因此我的工作方式和生活方式也不同于其他人。我每天都会花费大量的时间和精力，努力为客户提供最好的法律服务和建议。当然，我也有自己的休闲时间，我会选择去探索新的城市或者旅游新的国家，体验不同的文化和风俗。</s> \\
        \textcolor{blue}{\textbf{用户}: 那你是白天活跃还是晚上活跃啊？} \\
        \textbf{模型}: 作为一名律师，我的工作时间通常是在正常的工作时间内进行，即从早晨9点开始到下午6点结束。但是，如果有特殊情况需要，我也会在夜间或者周末进行工作，以便更好地为客户提供服务。当然，我也有自己的休闲时间，我会选择去探索新的城市或者旅游新的国家，体验不同的文化和风俗。</s> \\
        \bottomrule
    \end{tabular}
\end{table*}
\end{CJK*}
\subsection{Chinese Medical Model}
We used continued training to fine-tune the model on cMedQA2\footnote{https://github.com/zhangsheng93/cMedQA2}, resulting in a model that possesses both general Chinese instruction-following ability and expertise in Chinese medical question answering.

\paragraph{DataSets Overview} cMedQA 2.0 is the second version of the cMedQA dataset, designed for Chinese community-based medical question answering.

\paragraph{Details and Performance}
To preserve the original instruction-following capability, we fixed the format for medical instruction fine-tuning as follows: 
\texttt{\{"instruction": "Play the role of a professional doctor and answer the patient's questions.", "input": "Patient: xxx", "output": "1. xxxx.$\backslash$n2. xxx..."\}}.

Specifically, we continued fine-tuning from a Vicuna model fine-tuned for one epoch. This approach yielded a more specialized performance on medical Q\&A data compared to fine-tuning for three epochs. Additionally, due to the nature of the dataset, the model's responses became more standardized, often including phrases like "Consult a professional hospital for examination," etc. The model can be accessed here: \url{https://huggingface.co/Chinese-Vicuna/Chinese-Vicuna-continue-finetune-7epoch-cMedQA2}.

The training parameters and process are consistent with those used for the instruction-tuned model.
The corresponding performance is shown as Table~\ref{tab:med}.
We observed that the model provides more professional answers to many medical queries, showing a stronger medical question-answering capability compared to models that have not been fine-tuned to medical data. At the same time, since the medical model was developed through continued training, it retains strong instruction-following abilities in other general domains as well.

\subsection{Chinese Legal Model}
For the legal domain, we continued training the ChatV1 model using the \textbf{continue training} approach for nearly 6 epochs. The training data includes the Lawyer-LLaMA dataset~\cite{huang2023lawyer} and the CAIL dataset\footnote{https://github.com/thunlp/CAIL}.  
The fine-tuned model is available at: \url{https://huggingface.co/Chinese-Vicuna/Chinese-Vicuna-7b-legal-lora}.  

\paragraph{DataSets Overview} The Lawyer-LLaMA dataset~\cite{huang2023lawyer} is a publicly available Chinese legal dataset. The CAIL dataset consists of legal documents from criminal cases published on Chinese Judgment Online.

\paragraph{Details and Performance}
The training parameters and process are consistent with those used for the instruction-tuned model.

The results are presented in Table~\ref{tab:legal} and Table~\ref{tab:legal_general}. As shown, our legal model performs well across various legal question-answering tasks and also demonstrates a certain level of general-purpose capability. However, its ability to perform role-playing in general scenarios is somewhat limited, as the model tends to default to the role of a lawyer. This behavior is likely influenced by the extensive training on legal-specific tasks.



 
\FloatBarrier
\section{Conclusions} 
Chinese-Vicuna demonstrates that parameter-efficient fine-tuning can adapt LLaMA to Chinese tasks while maintaining hardware accessibility. Its domain-specific workflows in healthcare and law, coupled with open-source tools for quantization and deployment, offer a scalable solution for low-resource scenarios. Future work will integrate RLHF for alignment and dynamic knowledge retrieval to address temporal data gaps. By fostering community collaboration, Chinese-Vicuna aims to democratize Chinese LLM development across academic and industrial domains. In addition to this efficient approach, we also provide a full-parameter fine-tuning implementation for scenarios demanding higher performance, which is available at \url{https://github.com/LZY-the-boys/CustomLLMFinetuningHandbook}.

\bibliographystyle{plainnat}
\bibliography{neurips_2024}

\end{document}